\documentclass[conference]{IEEEtran}
\usepackage[left=0.635in,right=0.635in,top=0.75in,bottom=1.05in]{geometry}

\usepackage[numbers]{natbib}
\usepackage{hyperref}
\hypersetup{
    colorlinks=false,
    linkcolor=blue,
    citecolor=blue,
    urlcolor=blue,
}
\usepackage{caption}
\usepackage{subcaption}
\usepackage{multirow}
\usepackage{tabularray}
\usepackage{algorithmic}
\usepackage{graphicx}
\usepackage{textcomp}
\usepackage{fancyhdr}
\usepackage[table]{xcolor} 
\usepackage{colortbl}
\usepackage{makecell}
\usepackage{tabularx}
\usepackage{float}
\usepackage{amsmath}
\usepackage{amssymb}
\usepackage{amsfonts}
\usepackage{listings}
\usepackage{xcolor}
\usepackage{booktabs}
\usepackage[most]{tcolorbox}
\usepackage{subcaption}
\usepackage{tikz}
\usepackage{caption}
\usepackage{enumitem}
\usepackage{subcaption} 
\definecolor{PineGreenDarkest}{HTML}{4C9E8F}
\definecolor{PineGreenDark}{HTML}{70B3A1}
\definecolor{PineGreenMedium}{HTML}{94C8B4}
\definecolor{PineGreenLight}{HTML}{B8DDC6}
\definecolor{PineGreenLighter}{HTML}{D1EED9}
\definecolor{darkgreen}{rgb}{0.0, 0.5, 0.0} 
\lstdefinestyle{mypython}{
    language=Python,
    basicstyle=\ttfamily\small,
    keywordstyle=\color{blue}\bfseries,
    stringstyle=\color{darkgreen},
    commentstyle=\color{gray}\itshape,
    morekeywords={BaseModel, Field, List},
    columns=flexible,
    keepspaces=true,
    breaklines=true,
    showstringspaces=false,
    frame=single,
    rulecolor=\color{black},
    stepnumber=1,
    numbersep=5pt,
    xleftmargin=0.05\columnwidth,   
    xrightmargin=0.05\columnwidth,  
}

\begin{document}
%
\title{An Integrated Approach to AI-Generated Content in e-health}

\author{
    \IEEEauthorblockN{Tasnim Ahmed, 
    Salimur Choudhury}
    
    \IEEEauthorblockA{School of Computing,
    Queen's University, Ontario, Canada}

    \IEEEauthorblockA{\{tasnim.ahmed, s.choudhury\}@queensu.ca}
}

\maketitle

\begin{abstract}
Artificial Intelligence-Generated Content, a subset of Generative Artificial Intelligence, holds significant potential for advancing the e-health sector by generating diverse forms of data. In this paper, we propose an end-to-end class-conditioned framework that addresses the challenge of data scarcity in health applications by generating synthetic medical images and text data, evaluating on practical applications such as retinopathy detection, skin infections and mental health assessments. Our framework integrates Diffusion and Large Language Models (LLMs) to generate data that closely match real-world patterns, which is essential for improving downstream task performance and model robustness in e-health applications.
Experimental results demonstrate that the synthetic images produced by the proposed diffusion model outperform traditional GAN architectures. Similarly, in the text modality, data generated by uncensored LLM achieves significantly better alignment with real-world data than censored models in replicating the authentic tone.

\end{abstract}
\begin{IEEEkeywords}
Generative Artificial Intelligence,
e-health,
Synthetic Data,
Large Language Models, Diffusion
\end{IEEEkeywords}

\IEEEpeerreviewmaketitle

\section{Introduction}
Artificial Intelligence-Generated Content (AIGC) is transforming healthcare by revolutionizing data generation, usage, and analysis. Advances in Generative AI (GenAI) now allow synthetic data—like medical images and clinical text—to supplement real-world datasets. Access to real healthcare data remains limited due to strict privacy regulations, complex de-identification, and fragmented storage \cite{doi:10.1148/radiol.2020192224}. Synthetic data overcomes these barriers by providing diverse, privacy-compliant datasets, and supporting e-health applications like remote diagnostics, telemedicine, and clinical decision support within ethical and regulatory standards. Among e-health data modalities, images and text are the most extensively used and critically important in AI applications.
Approaches for synthetic medical image generation include Generative Adversarial Networks (GANs), variational autoencoders, and diffusion models. Recent studies show that GANs and diffusion models have been used to generate synthetic MRI for segmentation, enhance CT scans with synthetic pathologies, create de-identified chest radiographs, and synthesize contrast-enhanced CT images to reduce agent usage \cite{doi:10.1148/radiol.232471}. Diffusion provides an advantage over GANs by producing images with higher fidelity and diversity, though at the expense of generation speed \cite{NEURIPS2020_4c5bcfec}. Further exploration of class-conditioned diffusion models is essential for e-health that will allow disease-specific image generation and improve accuracy in tasks like classification, segmentation, and diagnostic support, where class-specific details are vital. Recent approaches to synthetic medical text generation combine Large Language Models (LLMs) and knowledge graphs with advanced NLP techniques \cite{10.1007/978-3-031-70381-2_14}. Key applications include clinical note generation, named entity recognition, and text classification. Studies demonstrate that models like GPT-4 and LLaMA produce accurate, diverse clinical text, reducing hallucinations and enhancing downstream tasks. However, due to ethical constraints, aligned and censored LLMs struggle to generate realistic medical text data, as sensitive information in real medical texts is difficult to replicate while maintaining privacy. To address this, there is a need to explore uncensored and unaligned LLMs in a controlled environment and develop more robust evaluation methods to better identify discrepancies between real and synthetic data.

To this end, to mitigate the research gap in generating synthetic image and text modality data, we propose two GenAI methods. Therefore, primary contributions of this research are: (1) a novel class-conditioned framework introducing a diffusion model for synthetic medical image generation and an uncensored LLM-based approach for generating synthetic medical text, addressing data scarcity in e-health applications; (2) the diffusion architecture outperforms traditional GANs, and the adoption of object parsing through structured output refines responses from the uncensored LLM; and (3) a robust, four-stage evaluation strategy for generated text data to enhance transparency in performance differences between real and synthetic data and to identify specific areas of improvement. We will make our framework and code publicly available upon manuscript review (github.com/tasnim7ahmed/gen-health).

\section{Methodology}
\begin{figure*}[h!]
    \centering
    \includegraphics[width=0.8\textwidth]{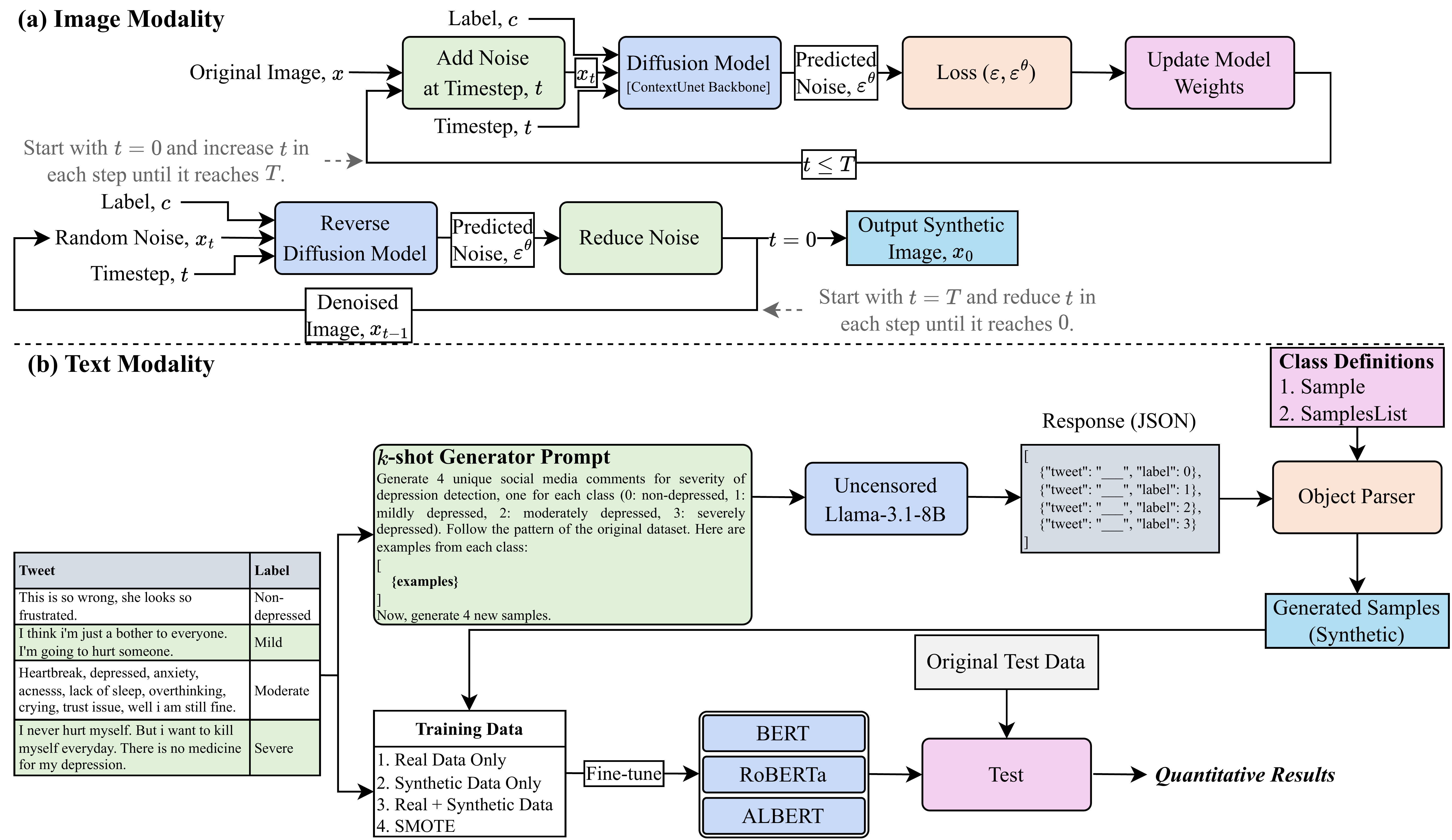}
    \caption{Proposed framework for training and evaluation of AI-generated e-health data. This framework generates both image and text data that approximates original samples, although the text modality does not require task-specific training. The generated data can be applied in diverse e-health contexts, including diagnosis, monitoring, personalized treatment, and remote care.}
    \label{fig:pipeline}
\end{figure*}

\figureautorefname~\ref{fig:pipeline} illustrates our proposed pipeline for the generation and evaluation of e-health data, focusing on the image and text modalities. A key feature of this framework is its class-conditioned design for both modalities, allowing for more controlled data generation.

\subsection{Image Modality}

The proposed model utilizes a class-conditioned diffusion approach for generating synthetic medical images, integrating classifier-free guidance \cite{ho2021classifierfree}
to enhance sample quality while maintaining generative flexibility.
Building on foundational U-Net architecture \cite{10.1007/978-3-319-24574-4_28}, a modified `ContextUnet' is implemented that incorporates class-conditioning elements to precisely control image generation.
The training begins with an initial input image, \( x \), a class label, \( c \), and a designated timestep, \( t \).
The `ContextUnet' processes image data through both downsampling and upsampling paths using convolutional and transposed convolutional layers with residual connections, providing,
\(
\text{Output} = \frac{x + \text{Conv}(\text{Conv}(x))}{\sqrt{2}}
\). Here, each `Conv' includes convolution, batch normalization, and GELU activation
for efficient feature extraction while preserving contextual information via skip connections. Image synthesis is conditioned on class labels through embeddings generated by \(
\mathbf{c}_{\text{emb}} = \text{GELU}(\mathbf{W}_2 \, \text{GELU}(\mathbf{W}_1 \, \mathbf{c}))
\). Here, \( \mathbf{W}_1 \) and \( \mathbf{W}_2 \) are learned weight matrices, and \( \mathbf{c} \) represents the class vector. These embeddings, combined with temporal embeddings, are integrated into the latent space and modulated through context masking,
\(
\mathbf{c} = \mathbf{c} \odot M \quad \text{where} \quad M \in \{0, 1\}^{n_{\text{classes}}}.
\)
Context masking allows partial conditioning and stochastic variations for diverse outputs within class constraints.
At each timestep, random noise $\epsilon$ is introduced to the image, producing a progressively noisier version $x_t$ that the model will learn to reconstruct.
The diffusion process follows a denoising schedule that provides a controlled transition from noise to coherent images. In the forward diffusion step, noise \( \epsilon \sim \mathcal{N}(0, I) \) is added to the input image \( x_0 \) as 
\(
x_t = \sqrt{\bar{\alpha}_t} \, x_0 + \sqrt{1 - \bar{\alpha}_t} \, \epsilon
\),
where \( \bar{\alpha}_t = \prod_{s=1}^{t} \alpha_s \).
`ContextUnet' then predicts this added noise $\epsilon^\theta$ which allows the model to estimate the noise level and adjust its parameters for better noise prediction accuracy.
During reverse diffusion, the model predicts the noise component \( \epsilon_\theta(x_t, c, t) \) to reconstruct the image, 
\(
x_{t-1} = \frac{1}{\sqrt{\alpha_t}} \left( x_t - \frac{1 - \alpha_t}{\sqrt{1 - \bar{\alpha}_t}} \epsilon_\theta(x_t, c, t) \right) + \sqrt{\beta_t} \, z
\),
with \( \beta_t \) representing the noise added at each timestep and \( z \sim \mathcal{N}(0, I) \). The classifier-free guidance mechanism aggregates conditional and unconditional noise predictions by:
\(
\tilde{\epsilon}_\theta(x_t, c, t) = (1 + w) \, \epsilon_\theta(x_t, c, t) - w \, \epsilon_\theta(x_t, \emptyset, t)
\), where \( w \) controls the conditioning strength to balance the fidelity and diversity of the generated images.
Depending on the noise prediction, a loss is calculated, allowing the weights to be updated for increased accuracy in denoising at each step.
The training objective involves denoising score matching, minimizing the mean squared error between the predicted and actual noise:
\(
\mathcal{L}(\theta) = \mathbb{E}_{x_0, \epsilon, t, c} \left[ \left\| \epsilon - \epsilon_\theta(x_t, c, t) \right\|^2 \right]
\).
In the generation phase, a random noisy image, $x_T$, is produced and iteratively denoised by the reverse diffusion model, progressively reducing noise until the final image, $x_0$, is generated, matching the class label $c$.
Additionally, class conditioning is improved by randomly applying a context mask \( M \), thereby generating conditional images without requiring an auxiliary classifier. The quality of the synthesized images is evaluated using Fréchet Inception Distance (FID), Structural Similarity Index Measure (SSIM), and Peak Signal-to-Noise Ratio (PSNR).

\subsection{Text Modality}
Censored and uncensored LLMs differ in content moderation and training methodologies. Censored LLMs go through stringent filters and alignment strategies, including supervised fine-tuning with curated datasets and safety layers, to prevent the generation of sensitive, harmful, or inappropriate content. On the other hand, uncensored LLMs do not have these restrictions, which allows them to generate more varied and open-ended text.
Using uncensored LLMs to generate synthetic medical data is crucial despite these inherent risks. Censored models may unintentionally suppress or alter subtle expressions and complex emotions, leading to less authentic and biased samples. Uncensored LLMs allow the creation of more genuine and diverse data, essential for developing reliable medical models. Furthermore, in fields requiring an accurate understanding of sensitive conditions, generating unfiltered data ensures that synthetic datasets reflect the true diversity and complexity of real-world scenarios.

In this study, the uncensored version of the Llama-3.1-8B \cite{dubey2024llama3herdmodels} is used to generate synthetic samples for a multi-class mental health text classification dataset. Implementing a $k$-shot prompting, where $k$ corresponds to the number of classes in the dataset, the generation process started with an instruction prompt that included one randomly sampled example from each class. Subsequently, the LLM was instructed to produce one new sample per class based on these initial examples. To address the alignment issues inherent in uncensored LLMs and facilitate automated parsing, we adopted a structured output generation strategy. This involved defining specific classes for object parsing from the response as shown below:

\begin{lstlisting}[style=mypython]
class Sample(BaseModel):
    tweet: str = Field(
        description="A social media comment extracted from Twitter")
    target: int = Field(
        description=(
            "Depression level: "
            "0 = non-depressed, "
            "1 = mildly depressed, "
            "2 = moderately depressed, "
            "3 = severely depressed"
        )
    )
class GeneratedSamples(BaseModel):
    samples: List[Sample]
\end{lstlisting}
Each LLM response generates a `GeneratedSamples' object that includes one sample from each of the $k$ classes. The proposed generation strategy allowed the LLM outputs to be parsed consistently, resulting in a balanced dataset across all classes.
Since the proposed approach utilizes a structured uncensored LLM for generating data samples, it can be incorporated to produce tabular data through modifications to the model classes. This involves adjusting the class attributes to align with specific database requirements, such as defining appropriate field types—including text, numerical, and binary—and providing detailed descriptions for each field.

The quality of generated data is evaluated through an experimental framework proposed by Schmidhuber et al. \cite{kruschwitz-schmidhuber-2024-llm}. The evaluation pipeline includes four training configurations: real data only, a composite of real and synthetic data, synthetic data only, and a Synthetic Minority Over-sampling Technique (SMOTE)-like method for class balancing. The objective is to investigate the impact of synthetic data in varied settings by fine-tuning text classifiers and evaluating their performance on downstream tasks. Transformer-based masked language models---BERT \cite{devlin-etal-2019-bert}, RoBERTa \cite{liu2019robertarobustlyoptimizedbert}, and ALBERT \cite{Lan2020ALBERT:}---are implemented, with performance measured through accuracy, F1 score, precision, and recall. The experiment design details are as follows.

\subsubsection{Real Data (Original)}
The real, original dataset \( D_{orig} \) is split into training (\( D_{orig-train} \)) and testing sets (\( D_{orig-test} \)). This initial experiment involves training classifiers on \( D_{orig-train} \) and validating their performance on \( D_{orig-test} \). The classifiers trained on real data set the benchmark against which synthetic and combined training scenarios are evaluated. This determines if the addition of synthetic data contributes positively to performance or introduces potential drawbacks.

\subsubsection{Composite}
This experiment evaluates the performance of classifiers trained on a combination of real and synthetic data. The training data comprises both \( D_{orig-train} \) and the generated synthetic data \( D_{synth} \) and can be expressed as: \(
D_{comp-train} = D_{orig-train} \cup D_{synth}
\). The idea is to see if the presence of synthetic data can enhance the generalizability and robustness of the model. The classifiers trained on these datasets are then evaluated on \( D_{orig-test} \).

\subsubsection{Synthetic}
The training set \( D_{synth} \) is used to fine-tune classifiers, which are then validated on \( D_{orig-test} \).  The synthetic-only training is validated through $5$-fold cross-validation to evaluate the stability and reliability of model performance across different splits. This experiment decides whether generated data can effectively replace real data by evaluating if models trained on artificial examples can generalize successfully to real-world data.

\subsubsection{SMOTE}
SMOTE mitigates class imbalance by incorporating synthetic samples into the minority class of \( D_{orig-train} \). This method adjusts the training set lengths such that:
\(
D_{comp-1} = D_{orig-1} + D_{synth-1}
\),
where \( D_{orig-1} \) represents the minority class in the original training set, and \( D_{synth-1} \) consists of synthetic samples for that class. The modified training set is then used to train classifiers, which are evaluated on \( D_{orig-test} \). This experiment tests whether synthetic oversampling can effectively balance classes and improve performance.

\section{Result and Discussion}
\subsection{Datasets}
\subsubsection{Diabetic Retinopathy \cite{Karthik2019APTOS}} The APTOS 2019 Blindness Detection dataset consists of retina scan images intended for the detection and classification of diabetic retinopathy. The dataset is organized by the severity of diabetic retinopathy, with images distributed across five classes: no diabetic retinopathy, mild, moderate, severe, and proliferate diabetic retinopathy.
\subsubsection{DermNet \cite{app10072488}} The DermNet dataset consists of approximately $23000$ images representing 23 distinct skin disease categories, sourced from the DermNet online portal. The dataset includes a diverse range of conditions such as acne, melanoma, eczema, psoriasis, and vascular tumours.

\subsubsection{DEPTWEET \cite{KABIR2023107503}} The DEPTWEET dataset comprises crowdsourced tweets labelled with depression severity levels. This dataset is founded on a typology based on psychological theory, designed to detect varying degrees of depression. Labels follow a four-level classification: Non-depressed, Mild, Moderate, and Severe. The labelling process utilizes the DSM-5 \cite{american2013diagnostic} diagnostic criteria and the Patient Health Questionnaire (PHQ-9) \cite{Kroenke2001606}, which identify nine symptoms linked to depression severity, including suicidal thoughts, lack of interest, and sleep disorders. The dataset presents a notable class imbalance, heavily skewed toward the Non-depressed class, with fewer samples in the Severe category.

\subsection{Image Modality}
\begin{figure}[h!]
    \centering
    \includegraphics[width=0.8\columnwidth]{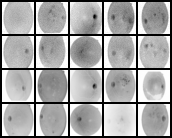}
    \caption{Class-conditioned diffusion model-generated synthetic diabetic retinopathy images (grayscale). The columns from left to right correspond to class labels: No\_DR, Mild, Moderate, Severe, and Proliferate\_DR. The first two rows display synthetic images generated by the model, while the last two rows contain real images for comparison.}
    \label{fig:synthetic_dr_images}
\end{figure}

\begin{figure*}[h!]
    \centering
    \includegraphics[width=0.9\textwidth]{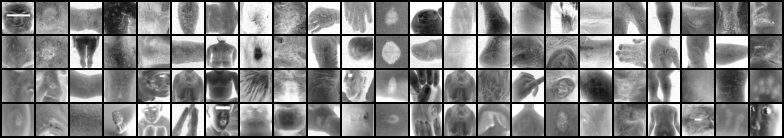}
    \caption{Class-conditioned diffusion model-generated synthetic images (grayscale) related to 21 dermatological conditions. The columns from left to right correspond to class labels. The first two rows display synthetic images generated by the model, while the last two rows contain real images for comparison.}
    \label{fig:synthetic_dermnet_images}
\end{figure*}

\begin{figure*}[h!]
    \centering
    \begin{subfigure}[t]{0.4\textwidth}
        \centering
        \includegraphics[width=\linewidth]{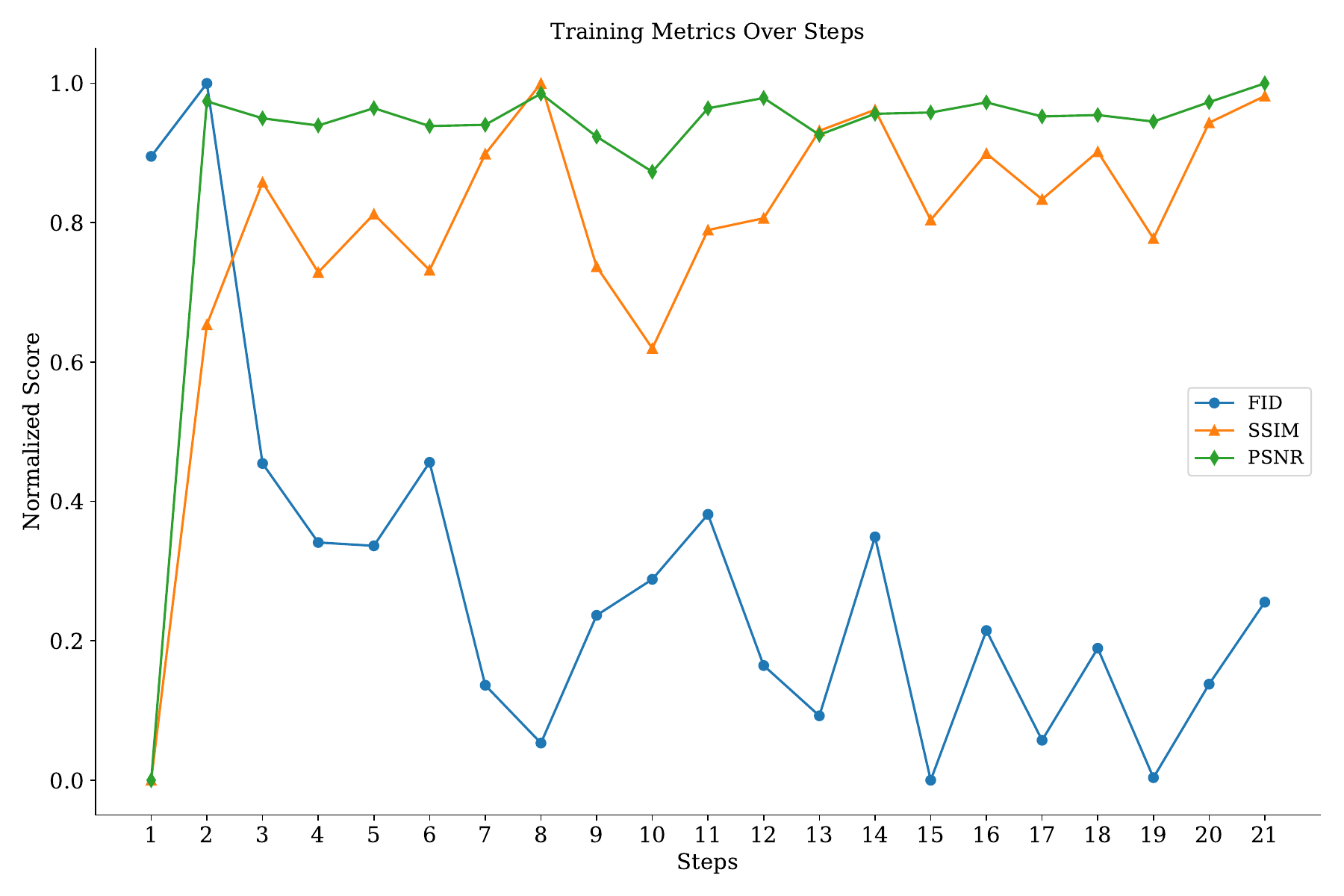}
        \label{fig:training_metrics_dr}
    \end{subfigure}
    \hfill
    \begin{subfigure}[t]{0.4\textwidth}
        \centering
        \includegraphics[width=\linewidth]{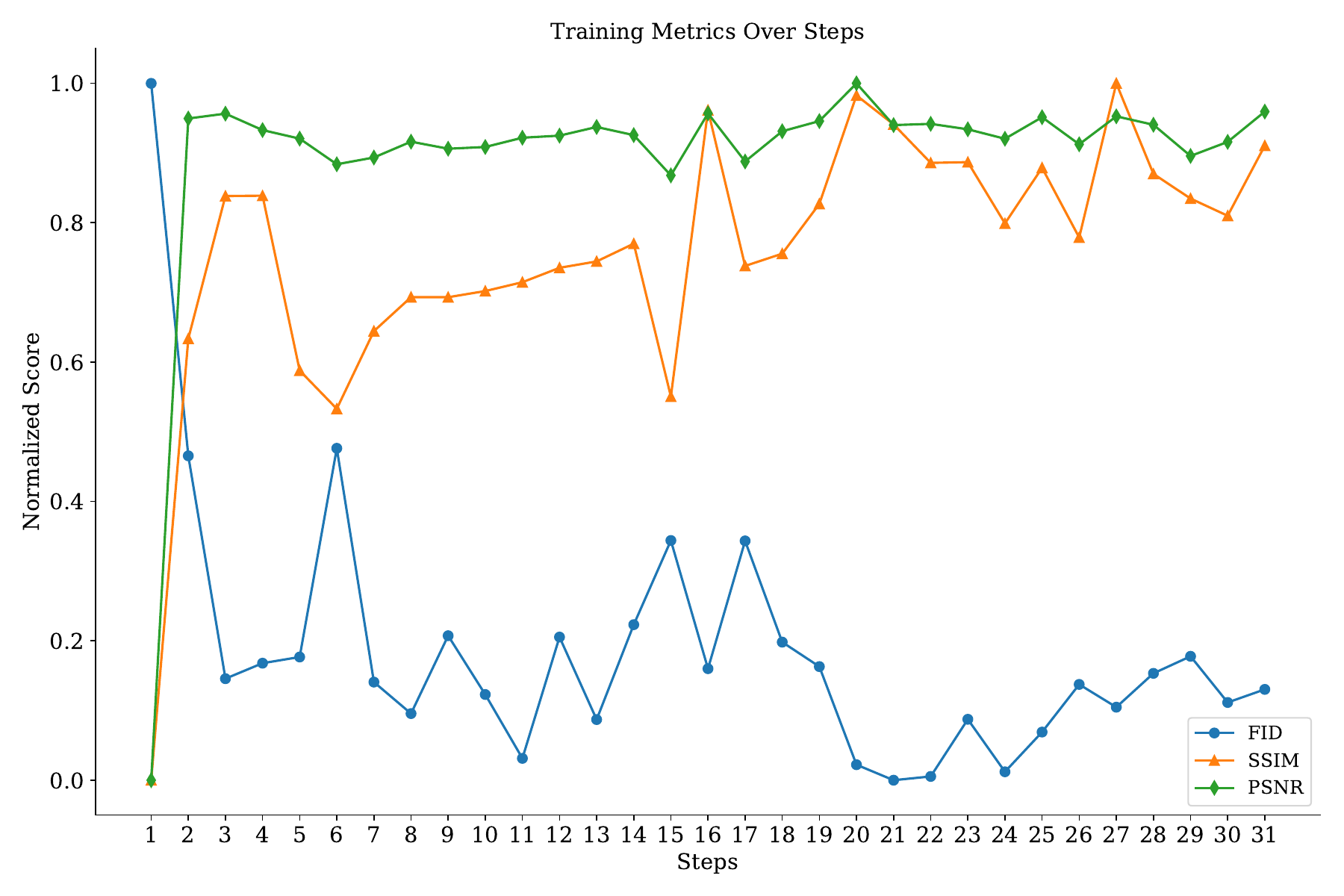}
        \label{fig:training_metrics_dermnet}
    \end{subfigure}
    \caption{Training metrics over steps for the diabetic retinopathy (left) and the dermnet (right) dataset. The normalized scores of FID, SSIM, and PSNR are shown. Each step corresponds to 50 epochs.}
    \label{fig:dual_training_metrics}
\end{figure*}

The proposed diffusion model was trained and evaluated on the DermNet and Diabetic Retinopathy datasets to provide a proof of concept, demonstrating that class-conditioned diffusion models can outperform leading GAN architectures. Our study empirically validates that the model can generate realistic synthetic images conditioned on class labels. We selected datasets with differing structural similarities: the Diabetic Retinopathy dataset features minimal inter-image differences, while the DermNet dataset includes higher structural variability within and between classes, representing various body parts. All samples were converted to grayscale for two reasons: (1) to reduce computational complexity by reducing the number of channels from three to one, and (2) because prior research shows grayscale images retain sufficient visual features for classification in similar tasks.
The proposed model is evaluated for $32\times32$ class-conditioned image generation. Table~\ref{tab:generative_metrics} shows that it outperforms Conditional GAN (CGAN) \cite{mirza2014conditionalgenerativeadversarialnets} and Auxiliary Classifier GAN (ACGAN) \cite{10.5555/3540261.3542061}, achieving significantly lower FID and higher SSIM and PSNR metrics. Although these results are promising, further improvement is possible, as an FID below $100$, SSIM above $0.5$, and PSNR near $20$ would closely approximate real distribution in synthetic images.
For SSIM, the metrics reveal that while FID and PSNR scores are similar across datasets, SSIM for retinopathy nears $0.5$, but is notably lower ($0.2097$) for DermNet in the best-performing model. Manual inspection suggests that this discrepancy may be due to structural variability in DermNet images, as images within the same class may represent different body parts, leading to lower SSIM scores despite the class similarity.
Two hyperparameters notably impact the model performance: guidance strength $w$ and sampling steps $T$. Optimal FID was achieved with $w=2.0$ after experimenting with values from $0.5$ to $4.0$ in $0.5$ intervals. Empirical and theoretical analysis shows $T$ significantly affects quality, with higher $T$ improving quality but slowing sampling. Setting $T=400$ balanced sampling quality and speed effectively across $w$ values. Figures~\ref{fig:synthetic_dr_images} and \ref{fig:synthetic_dermnet_images} display real and generated images, and Figure~\ref{fig:dual_training_metrics} shows metric trends over training epochs.

\begin{table}[ht]
    \centering
    \caption{Experimental Results on Medical Image Generation}
    \label{tab:generative_metrics}
    \begin{tabular}{llccc}
        \toprule
        \textbf{Dataset} & \textbf{Model Name} & \textbf{FID $\downarrow$} & \textbf{SSIM $\uparrow$} & \textbf{PSNR $\uparrow$} \\
        \midrule
        \multirow{3}{*}{Dermnet}
            & CGAN & 266.7141 & 0.1644 & 9.7041 \\
            & ACGAN & 239.8069 & 0.1683 & 10.6513 \\
            & Diffusion (Proposed) & 203.1043 & 0.2097 & 10.8449 \\
        \midrule
        \multirow{3}{*}{Retinopathy}
            & CGAN & 339.0033 & 0.4800 & 16.7719 \\
            & ACGAN & 300.3846 & 0.3470 & 14.4552 \\
            & Diffusion (Proposed) & 245.7854 & 0.5593 & 17.8413 \\
        \bottomrule
    \end{tabular}
\end{table}

\subsection{Text Modality}
Four experimental configurations were evaluated: Original, Composite, Synthetic, and SMOTE, with the effectiveness of censored and uncensored (proposed) LLMs evaluated across each setting (Table~\ref{tab:evaluation_metrics}). As expected, RoBERTa generally outperformed BERT in most experiments, given its significantly larger parameter size. Interestingly, however, despite being much smaller in parameter size than both BERT and RoBERTa, ALBERT achieved comparable performance in the classification task. Comparative analysis shows that classifiers trained using synthetic data generated by the uncensored Llama-3.1-8B consistently outperformed those trained on censored data, achieving improved evaluation scores across all experiments. The performance gap between censored and uncensored models can be attributed to the nature of the text samples they generate. As shown in Table~\ref{tab:llm_samples}, censored LLM produce text with a more formal, sanitized tone, exemplified by phrases like ``Feeling hurt and frustrated," which appear polished and restrained. This tone lacks the raw, authentic quality typical of social media comments. In contrast, the uncensored LLM samples more closely mirror the unfiltered, personal tone of social media posts, such as ``I've been struggling with suicidal thoughts," providing a candid portrayal of mental states. Examination of samples from censored and uncensored LLM reveals that for moderate or severely depressed classes, the censored LLM generated responses with a less negative tone, avoiding sensitive words like `suicide' or `kill.' This occurs because aligned LLMs are trained to filter out such content, resulting in generated samples that more closely resemble non-depressed or mildly depressed categories, despite being labelled as moderate or severe. This discrepancy contributes to the reduced performance of the censored LLM, as the generated data lacks the authentic severity seen in real-world samples.

Experiment types are crucial in highlighting the need for uncensored LLMs for such tasks. In the composite experiment, adding censored LLM-generated data to the original training set decreases classification performance, revealing a significant mismatch in data patterns. Conversely, when combined with uncensored LLM data, classifiers perform better than with original data alone, indicating that uncensored synthetic data closely replicates original data patterns. Similar trends appear in synthetic experiments, where classifiers trained with censored data perform poorly. In SMOTE, balancing data with censored samples even lowers performance below the original data, while using synthetic data from our framework significantly improves performance. This is crucial in e-health, where imbalanced data often biases models. Empirical results show that our framework effectively addresses this issue. It is observed that generating an equal number of responses from uncensored LLM took slightly more time than from censored ones. This is because unaligned LLMs generally follow instructions less accurately than aligned models, leading to occasional failures or filtering out of responses due to the structured response constraints set by sample class definition and object parser.

\begin{table}[ht]
    \centering
    \caption{Evaluation Metrics for BERT, RoBERTa, and ALBERT Models Across Different Experiments and LLM types}
    \label{tab:evaluation_metrics}
    \scriptsize
    \begin{tabular}{lclcccc}
        \toprule
        \textbf{Exp.} & \textbf{Censored?} & \textbf{Model} & \textbf{Acc. $\uparrow$} & \textbf{Prec. $\uparrow$} & \textbf{Recall $\uparrow$} & \textbf{F1 $\uparrow$} \\
        \midrule
        \multirow{3}{*}{Original} 
            & \multirow{3}{*}{---} 
                & BERT & 0.825 & 0.467 & 0.495 & 0.458 \\
            &  & RoBERTa & 0.848 & 0.490 & 0.492 & 0.466 \\
            &  & ALBERT & 0.833 & 0.569 & 0.470 & 0.475 \\
        \midrule
        \multirow{6}{*}{Composite} 
            & \multirow{3}{*}{\checkmark} 
                & BERT & 0.802 & 0.460 & 0.358 & 0.379 \\
            &  & RoBERTa & 0.812 & 0.462 & 0.560 & 0.486 \\
            &  & ALBERT & 0.815 & 0.459 & 0.487 & 0.466 \\
        \cmidrule{2-7}
            & \multirow{3}{*}{$\times$} 
                & BERT & 0.833 & 0.478 & 0.523 & 0.477 \\
            &  & RoBERTa & 0.853 & 0.523 & 0.550 & 0.517 \\
            &  & ALBERT & 0.815 & 0.510 & 0.612 & 0.538 \\
        \midrule
        \multirow{6}{*}{Synthetic} 
            & \multirow{3}{*}{\checkmark} 
                & BERT & 0.712 & 0.360 & 0.459 & 0.351 \\
            &  & RoBERTa & 0.777 & 0.409 & 0.510 & 0.421 \\
            &  & ALBERT & 0.718 & 0.323 & 0.410 & 0.355 \\
        \cmidrule{2-7}
            & \multirow{3}{*}{$\times$} 
                & BERT & 0.788 & 0.440 & 0.539 & 0.457 \\
            &  & RoBERTa & 0.818 & 0.438 & 0.513 & 0.447 \\
            &  & ALBERT & 0.758 & 0.361 & 0.461 & 0.379 \\
        \midrule
        \multirow{6}{*}{SMOTE} 
            & \multirow{3}{*}{\checkmark} 
                & BERT & 0.815 & 0.419 & 0.486 & 0.419 \\
            &  & RoBERTa & 0.807 & 0.362 & 0.424 & 0.342 \\
            &  & ALBERT & 0.722 & 0.420 & 0.363 & 0.355 \\
        \cmidrule{2-7}
            & \multirow{3}{*}{$\times$} 
                & BERT & 0.848 & 0.582 & 0.514 & 0.477 \\
            &  & RoBERTa & 0.823 & 0.452 & 0.502 & 0.445 \\
            &  & ALBERT & 0.838 & 0.451 & 0.374 & 0.398 \\
        \bottomrule
    \end{tabular}
\end{table}

\begin{table}[ht]
    \centering
    \caption{Samples from LLM generated data}
    \label{tab:llm_samples}
    \begin{tabular}{ll}
        \toprule
        \textbf{Llama-3.1-8B} & \textbf{Llama-3.1-8B (Uncensored)} \\
        \midrule
        \begin{minipage}[t]{0.45\columnwidth}
            \textbf{Sample 1:} ``I feel like I'm drowning in darkness. The thought of living another day is too much to bear. Please help me." (3) \\
            \textbf{Sample 2:} ``I just had a huge fight with my best friend. Feeling hurt and frustrated. Not sure how to move forward." (2)
        \end{minipage}
        &
        \begin{minipage}[t]{0.45\columnwidth}
            \textbf{Sample 1:} ``i've been struggling with suicidal thoughts for weeks now and i don't know how much longer i can keep going." (3)\\
            \textbf{Sample 2:} ``my therapist told me that my constant self-criticism is actually a coping mechanism for deeper emotional pain but it's hard to see that when all i can feel is numbness." (2)
        \end{minipage}
        \\
        \bottomrule
    \end{tabular}
\end{table}

\section{Limitations}
While the proposed framework demonstrates considerable potential in generating synthetic e-health data, several limitations should be acknowledged. Firstly, the study is constrained by the use of a limited number of datasets which may affect the generalizability of the results across a broader range of medical conditions. Additionally, the generation of low-resolution grayscale images may omit critical visual details essential for accurate medical diagnosis. Moreover, the current evaluation relies primarily on quantitative metrics, acknowledging the need for qualitative data analysis by medical domain experts or specialists to ensure the clinical relevance and practical utility of the synthetic data.

\section{Conclusion}
We developed an integrated framework for AIGC in the e-health domain, proposing class-conditioned diffusion models for synthetic medical images and an uncensored Llama-3.1-8B-based pipeline for medical text data. Our results demonstrate that the diffusion-based approach surpasses traditional GAN architectures in producing high-quality images, while text classifiers trained with our synthetic data exhibit improved performance, effectively addressing data scarcity in e-health applications. However, the use of uncensored LLMs carries the risk of generating sensitive or inappropriate content, which could have adverse effects if misused. Additionally, diffusion models have the potential to be exploited for creating harmful deepfakes, posing threats to the integrity of healthcare information. Therefore, it is essential to implement robust ethical guidelines and safeguards to ensure the responsible use of these technologies. Future work will focus on mitigating these risks (e.g., semi-alignment of uncensored LLMs by fine-tuning for specific downstream data generation tasks) and incorporating qualitative assessments from medical experts to validate the clinical relevance and safety of the generated content.
\bibliographystyle{IEEEtran}
\bibliography{ref}

\end{document}